# PROBLEM STRUCTURE AND EVIDENTIAL REASONING

*Richard M. Tong*
*Lee A. Appelbaum*

Advanced Decision Systems
201 San Antonio Circle, Suite 286
Mountain View, CA 94040.

## 1. INTRODUCTION

In our previous series of studies to investigate the role of evidential reasoning in the RUBRIC system for full-text document retrieval (Tong *et al.*, 1985; Tong and Shapiro, 1985; Tong and Appelbaum, 1987), we identified the important role that problem structure plays in the overall performance of the system. In this paper, we focus on these structural elements (which we now call "semantic structure") and show how explicit consideration of their properties reduces what previously were seen as difficult evidential reasoning problems to more tractable questions.

In the first part of the paper we discuss the nature of semantic structures in the RUBRIC system. We follow this with a discussion of the role that evidence plays. Then finally we show how certain evidential reasoning issues have a much clearer interpretation within the extended structures we have provided.

## 2. SEMANTIC STRUCTURES

To understand the nature of the retrieval problem, we start by supposing that there is a database of documents, denoted by $S$, in which the user is potentially interested. In response to a query, Q, the retrieval system returns the set of retrieved documents, denoted R, that are purported to be *relevant* to that particular query. In general, this set R will only be an approximation to the actual set of relevant documents, denoted $R^*$, contained in the database. The IR problem is thus one of designing a retrieval system that maximizes the intersection of R and $R^*$ (i.e., maximizing *recall*) whilst minimizing $R - R^*$ (i.e., maximizing *precision*).

The key issue here is what do we mean by the relevance of a document to a user's query. To understand this, we must distinguish between the subject matter of a document and the utility of the document to the user. So for example, a retrieved document might be about the topic of the user's query, but it might not be useful because the user has seen it before, or because it is a summary of a longer document, or any similar reason. Thus relevance includes a notion of usefulness for the task at hand. However, in the current version of RUBRIC we have no way of describing this idea of a user goal, and so we approximate by asserting that a document is relevant if its subject matter is the same as the subject of the user's query.

Our notion of relevance is further modified by the recognition that in most cases the decision as to whether a document is about the topic of a query cannot be made with absolute certainty. That is, a document can be about a topic to a certain degree; ranging from not about a topic at all, to definitely about a topic. Notice too, that a document can be about many topics. The information retrieval problem is thus one of asking the question "Is the document about topic X?" rather than the question "What is the document about?"

The central idea behind the RUBRIC knowledge representation is that a retrieval concept can be specified in terms of its constituent parts which thereby form a definition for that concept. In RUBRIC the user specifies the components of retrieval concepts by defining their "attributes." To illustrate, let us suppose that the user is interested in documents about meetings. Such events are composed of a number of elements of interest to the user, for example, the basic action (i.e., the act of meeting), the people involved, the topics of discussion, and the location of the meeting. The purpose of attributes, therefore, is to enable the user to define what it is about the main concept that is relevant for retrieval. We provide a single rule type for specifying attributes.

The ATTRIBUTE rule is used to capture the idea that concepts have components (or attributes), and that knowledge of these components may be used to help establish the presence of the concept itself. So for example if we take the attributes mentioned above, these would be defined in RUBRIC as:

(ATTRIBUTE **meetings** *action*)
(ATTRIBUTE **meetings** *actors*)
(ATTRIBUTE **meetings** *topic*)
(ATTRIBUTE **meetings** *location*)

where the syntax is LISP-like and we use emboldened text to indicate concepts and italicized text to indicate attribute names.

In conjunction with ATTRIBUTE rules we also need a mechanism for specifying the value of the attributes. In RUBRIC this is achieved by a rule of type DEFINES. So for example:

(DEFINES *location* (\*OR\* **moscow washington vienna geneva**))

where **moscow**, **washington**, **vienna**, and **geneva** are the concept names for various locations of interest, and \*OR\* denotes logical disjunction.

Another important feature of the RUBRIC knowledge representation is that it allows us to express taxonomic relationships between the retrieval concepts of interest. So, for example, if we are interested not only in the general class of **meetings**, but also in specific types of **meetings**, then we want to be able to express this in RUBRIC. Accordingly, we provide two rule types for expressing taxonomic relationships. (Notice that we are not as yet saying anything about direct evidence (i.e., the text itself) but are only concerned with the semantic structure of the domain.)

The first is the SUBSET rule. This rule type allows us to express the relationship between a sub-set of a set and the set itself. For example:

(SUBSET **meetings diplomatic-meetings**)

where again the syntax here is LISP-like, and the rule expresses the idea that **diplomatic-meetings** are a subset of all **meetings**.

Similarly, the INSTANCE rule allows us to express the relationship between an element of a set and the set itself. For example:

(INSTANCE **diplomatic-meeting us-soviet-summit**)

which expresses the idea that within the set of **diplomatic-meetings** we are specifically interested in any document about a **us-soviet-summit**.

Of course, this taxonomy can be made as complete as we wish simply by adding more rules. For example:

(INSTANCE **diplomatic-meetings sino-soviet-summit**)
(SUBSET **meetings white-house-meetings**)
(INSTANCE **white-house-meetings press-conference**)
(INSTANCE **white-house-meetings cabinet-meeting**)

would extend the taxonomy to include other kinds of **diplomatic-meetings** and also create a sub-taxonomy of **white-house-meetings**.

Concept attributes and their associated values are inherited from parent concepts unless specified locally. So for example, the location attribute for **white-house-meetings** would be defined as:

(DEFINES **white-house-meetings**:*location* **white-house**)

where the notation **concept**:*attribute-name* is used to distinguish variants of global attributes.

Obviously, with this representation we can build a detailed semantic description of interesting retrieval concepts. Notice that whilst the description can be used as a basis for the evidential reasoning we want to perform (see the next section), it could equally well be used to guide further knowledge engineering or be used to provide explanations.

## 3. EVIDENTIAL STRUCTURES

In the preceding section we described how the semantic structure of documents of interest can be encoded in RUBRIC. We now describe how to specify the textual evidence required to test whether the document under consideration matches the semantic structure. We first describe the basic string matching operations that RUBRIC can perform, and then describe the two types of inferential rule used to propagate the uncertainty values.

### 3.1. Text Reference Expressions

We provide a text reference language (henceforth TRL) that is used to describe patterns of text that should be searched for in support of retrieval concepts. The language consists of a number of basic operators that can be applied to one or more text arguments.

The first group of operators are just RUBRIC's equivalents of the conventional logical operators. The operators *AND* and *OR* can take multiple arguments, the operator *NOT* takes a single argument. So for example:

    (*OR* "KREMLIN" "MOSCOW" "RUSSIA")
    (*AND* "REAGAN" (*OR* "MOSCOW" "BEIJING"))
    (*NOT* (*AND* "REAGAN" "MOSCOW"))

are all legal expression in the TRL where upper-case quoted text is the actual pattern to matched in the body of the document.

The second group of operators are those we call distance operators. These take a pair of text arguments and return a value which represents the distance between them. The NEAR-W, NEAR-S and NEAR-P operators all return a value which is a normalized measure of the distance in words, sentences or paragraphs between their arguments. So for example:

    (NEAR-W "PRESIDENT" "REAGAN")

performs a test to see how close the words "PRESIDENT" and "REAGAN" are in the document.

The third main group of operators are those we call location operators. The SENTENCE and PARAGRAPH operators test to see if their arguments occur within the same sentence or paragraph in the document. The PRECEDES operator takes two keyword arguments and tests whether one occurs before the other. The WITHIN operator takes a numerical argument followed by two keyword arguments and tests whether they are within the specified number of words of one another. The PHRASE operator takes multiple text arguments and tests whether the phrase defined by concatenating the keywords occurs within the document. Examples are:

    (SENTENCE "GORBACHEV" "REAGAN")
    (PARAGRAPH "GORBACHEV" "GHANDI")
    (PRECEDES "SINO" "SOVIET")
    (WITHIN 10 "GORBACHEV" "REYKJAVIK")
    (PHRASE "STRATEGIC" "ARMS" "LIMITATION" "TALKS")

Several of the operators in the TRL can also take concepts as arguments. For example, all the logical operators, *AND*, *OR* and *NOT*, can be used in this way. In addition, RUBRIC has two non-traditional operators, BEST-OF and WEIGHT-OF, which take concepts as arguments and which capture the idea that, in the first case, any one of the arguments would be appropriate so we might as well take the best, and in the second that the more arguments that are true the better. RUBRIC also has a number of experimental features for general purpose synonym handling and for concept proximity testing.

### 3.2. Inferential Rules

To provide the necessary links between the text expressions and the concept taxonomies, the RUBRIC knowledge representation includes two inferential rules. These are the principal carriers of the uncertainty information which is appended to the rule as an uncertainty value. The RUBRIC system is designed to support a variety of uncertainty representations (e.g., standard probability, infinite-valued logics, various interval representations, and linguistic variables) and these can be selected as required.

The EVIDENCE rule is used to link text reference expressions to concepts. It captures the notion that text expressions are used as direct evidence in determining the relevance of the document to the retrieval topic. So for example:

(EVIDENCE **moscow** ((*OR* "MOSCOW" "KREMLIN") $\alpha$))

where "MOSCOW" and "KREMLIN" are text strings, and the value $\alpha$ is the degree of relevance to be assigned to the concept **moscow** if either of the test strings "MOSCOW" or "KREMLIN" are found in the document.

The IMPLIES rule is used for indirect evidence. That is, it is used to link retrieval concepts that are not in any taxonomic relationship to one another. So for our example domain of meetings, we might have:

(IMPLIES **salt** (**us-soviet-summit** $\beta$))

where **salt** is the concept name for the Strategic Arms Limitation Treaty. The value $\beta$ is the degree of relevance we wish to assign to a document that describes a **us-soviet-summit**, when we are in fact interested in documents about **salt**.

Our research to date has led us to conclude that the proper interpretation of these evidential relevance values is as the lower bound on our belief that the antecedent of the rule can be taken as an indicator of the consequent concept. That being the case, the default is for $\alpha$ and $\beta$ to be real numbers in the interval $[0,1]$. Part of our current research is designed to investigate whether the users of RUBRIC do indeed see the evidence values in this way.

## 4. EVIDENTIAL REASONING ISSUES

In this section we present a preliminary discussion of the evidential reasoning issues raised by our distinction between semantic and evidential structure. At the workshop itself we will present the results of some experiments which illustrate the advantages of this partitioning. We are especially interested in the internal form of the semantic structures and how these dictate the kinds of evidential operations we can perform.

By making a distinction between the semantics and the "syntax" of the document retrieval problem, we gain a more detailed understanding of the role that uncertainty plays in this domain. In particular, we see that uncertainty in the evidential structures is related to the degree of evidence or belief we wish to accumulate given the occurrence of certain patterns of text. Uncertainty in the semantic structures, on the other hand, is primarily concerned with issues such as the completeness and coherence of the basic structure, and also with the relative preferences amongst concepts when the structure is used for retrieval. In a system in which it is not possible to make these distinctions (e.g., the earlier version of RUBRIC) writing rules, assigning weights and defining uncertainty calculi becomes increasingly difficult as the rule-base gets larger and the various uncertainty concepts intermingle.

To illustrate, let us focus on the problem of combining evidence. The mechanism provided in RUBRIC for this is the COMBINE rule. Such rules have the following

syntax:

    (COMBINE **concept** *function-name*)

where **concept** is the concept for which we need to combine evidence, and *function-name* is the name of the actual combining function. The idea here is to allow the user to define arbitrary combining functions tailored to the special needs of the domain. We provide default combining functions, however, and in the case that the user does not specify a particular function, then one of these defaults is selected.

The first interesting case is the one in which we have several EVIDENCE and/or IMPLIES rules for a particular concept and need to specify how each contributes to the overall weight of evidence assigned to that concept. This is, of course, the problem we usually consider, and the key combining issues relate to the "independence" of the various sources of information. If we have no knowledge about this, then we can invoke the RUBRIC default which is to be as conservative as possible and simply take the maximum over all the inputs. That is:

$$v(c) = \max [\, v(c_i) \,]$$

where $v(c)$ denotes the resulting value for the concept and $v(c_i)$ denotes the values from the individual IMPLIES/EVIDENCE rules. Obviously, if we have more knowledge then we can adjust the combining function appropriately. Notice that within any given evidential structure we might have multiple combination functions, depending upon the assumptions we choose to make.

The second case of interest is the one in which we have propagated values through to the semantic structures and then need to specify how they should be combined in the subset/instance hierarchy. The intended semantics of our SUBSET/INSTANCE rules dictate that evidence for a subset, or an instance, is evidence for the larger class. However, since documents can be about multiple topics we interpret evidence for multiple subsets not as a conflict, but as confirmation that the larger topic is being discussed. The default is simply to take the maximum of the evidences, but obviously the RUBRIC system would allow for customized combination. So we have as a default:

$$v(c) = \max [\, v(s_i) \,]$$

where $v(s_i)$ denotes the contribution from the individual SUBSET/INSTANCE rules.

The third case is more complex, and is the one in which we wish to combine evidence from ATTRIBUTE rules. The main issue here is whether certain attributes are more indicative than others. Partly this is a function of personal preference, but partly it is a function of the concepts themselves. The RUBRIC system allows arbitrarily complex functions, but as a default we assume that the more evidence we have the better, and so:

$$v(c) = \frac{1}{N} \sum_{i=1}^{N} v(a_i)$$

where $v(a_i)$ denotes the contribution from the individual ATTRIBUTE rules.

One level of customization of attribute combining functions can be done by specifying whether attributes are either necessary, sufficient, or auxiliary in determining the value of their associated concept. So for example we might modify our attribute definitions for **us-soviet-summit** to be:

(ATTRIBUTE **us-soviet-summit** ((*NEC* *actors*) 0.6))
(ATTRIBUTE **us-soviet-summit** (*AUX* *location*))
(ATTRIBUTE **us-soviet-summit** (*AUX* *action*))

which expresses the idea that in determining the relevance of a document to a request for information about a **us-soviet-summit** it is necessary that the value of the *actors* attribute is at least 0.6 and that if it is then we can also take into account the values of the auxiliary attributes *location* and *action*. Functionally, this might look like:

if $v(actors) < 0.6$ then
   $v(\text{us-soviet-summit}) = 0.0$
otherwise
   $v(\text{us-soviet-summit}) = \min[\ 1.0,\ v(actors) + v(location) + v(action)\ ]$

Finally, some of the other evidential reasoning issues that we are currently exploring are: (1) measures of "goodness" for semantic structures and associated models of the interaction between goodness and belief, (2) models of user needs and their impact on the semantic and evidential structures, and (3) general purpose control strategies for reasoning over our extended knowledge representation.

## 5. SUMMARY

The discussion above is an attempt to provide support for our conjecture that the problem of evidential reasoning is one which is to be resolved at the semantic level rather than at the syntactic one. We believe that in any given problem it is imperative that we develop an understanding of the major reasoning patterns before we attempt any formal modelling of the uncertainty. In AI systems such as RUBRIC this means that the quality of the reasoning will often be dominated by a proper structuring of the knowledge and reasoning, rather than by questions about the formal properties of the uncertainty calculus. That is not to say, of course, that we should ignore the question of uncertainty representation and manipulation, but that there may be higher-level questions that have more overall significance and whose resolution may simplify the representational choices. We note that this need to focus on structure has been identified by other researchers (e.g., Toulmin, 1958; Cohen, 1985; Fox, 1986), and we submit that in our own efforts to address the characteristics of one specific application we have been able to demonstrate the value of this point of view.

The current version of RUBRIC is implemented in CommonLisp and C on a Sun Microsystems SUN-3 Workstation, and makes use of an ADS proprietary object-oriented programming environment (Cation, 1986). Within RUBRIC there are a number of modules that provide the user with facilities for creating and editing rules, for browsing existing rule-bases, for performing various kinds of sensitivity analysis, and for performing various kinds of retrieval analysis. Additional tools can be added easily as necessary, and the systems can be made to work with a variety of database management systems. The user interface is highly interactive and makes extensive use of graphics.